\pdfoutput=1

\documentclass[11pt]{article}

\usepackage[]{acl}

\usepackage{times}
\usepackage{latexsym}
\usepackage{booktabs}
\usepackage{multirow}
\usepackage{algorithmic}
\usepackage{tcolorbox}
\tcbuselibrary{skins}
\usepackage{xspace}
\usepackage{enumitem}
\usepackage{textcomp}
\usepackage{adjustbox}
\usepackage{wrapfig}
\usepackage{multirow}
\usepackage{amsthm}
\usepackage{subcaption}
\usepackage{comment}
\usepackage[normalem]{ulem}
\usepackage{fancyvrb}
\usepackage{hyperref} 
\usepackage{float}
\usepackage{tabularx} 
\usepackage{caption}
\usepackage{placeins}

\usepackage[T1]{fontenc}

\usepackage[utf8]{inputenc}

\usepackage{microtype}
\usepackage{tabularx}

\title{GATE X-E : A Challenge Set for Gender-Fair Translations from Weakly-Gendered Languages}

\author{Spencer Rarrick \And Ranjita Naik\thanks{All authors are affiliated with Microsoft.}
\thanks{Contact author at \texttt{\scriptsize ranjitan@microsoft.com}.} \And Sundar Poudel  \And     Vishal Chowdhary
}

\begin{document}
\maketitle
\begin{abstract}
 
Neural Machine Translation (NMT) continues to improve in quality and adoption, yet the inadvertent perpetuation of gender bias remains a significant concern. Despite numerous studies on gender bias in translations into English from weakly gendered-languages, there are no benchmarks for evaluating this phenomenon or for assessing mitigation strategies. To address this gap, we introduce GATE X-E, an extension to the GATE \citep{rarrick2023gate} corpus, that consists of human translations from Turkish, Hungarian, Finnish, and Persian into English. Each translation is accompanied by feminine, masculine, and neutral variants. The dataset, which contains between 1250 and 1850 instances for each of the four language pairs, features natural sentences with a wide range of sentence lengths and domains, challenging translation rewriters on various linguistic phenomena. Additionally, we present a translation gender rewriting solution built with GPT-4 and use GATE X-E to evaluate it. We open source our contributions to encourage further research on gender debiasing.

\end{abstract}

\section{Introduction}

Despite dramatic improvement in general NMT quality and breadth of supported languages over recent years \citep{nllbteam2022language}, gender bias in NMT output remains a significant problem \citep{piazzolla2023good}. One such type of gender bias is spurious gender-markings in NMT output when none were present in the source. This occurs most frequently when translating from a weakly-gendered language into a more strongly gendered one. We explore this phenomenon in translations from Turkish, Persian, Finnish, and Hungarian into English. 

Gender can be marked in English through gendered pronouns (\emph{he}, \emph{she}, etc.) and possessive determiners (\emph{his}, \emph{her}), or through a limited number of intrinsically gendered nouns (\emph{mother}, \emph{uncle}, \emph{widow}, etc), many of which are kinship terms. 

In each of our selected set of source languages, all personal pronouns are gender-neutral, such as \emph{O} in Turkish meaning \emph{he/she/singular they}. These languages do use some intrinsically gendered noun words, but not necessarily for all of the same concepts that English does. Turkish differentiates \emph{mother} (anne) from \emph{father} (baba), but does not differentiate \emph{nephew} from \emph{niece} (both are \emph{yeğen}).

\begin{figure}[t]
\centering \includegraphics[width=0.9\linewidth]{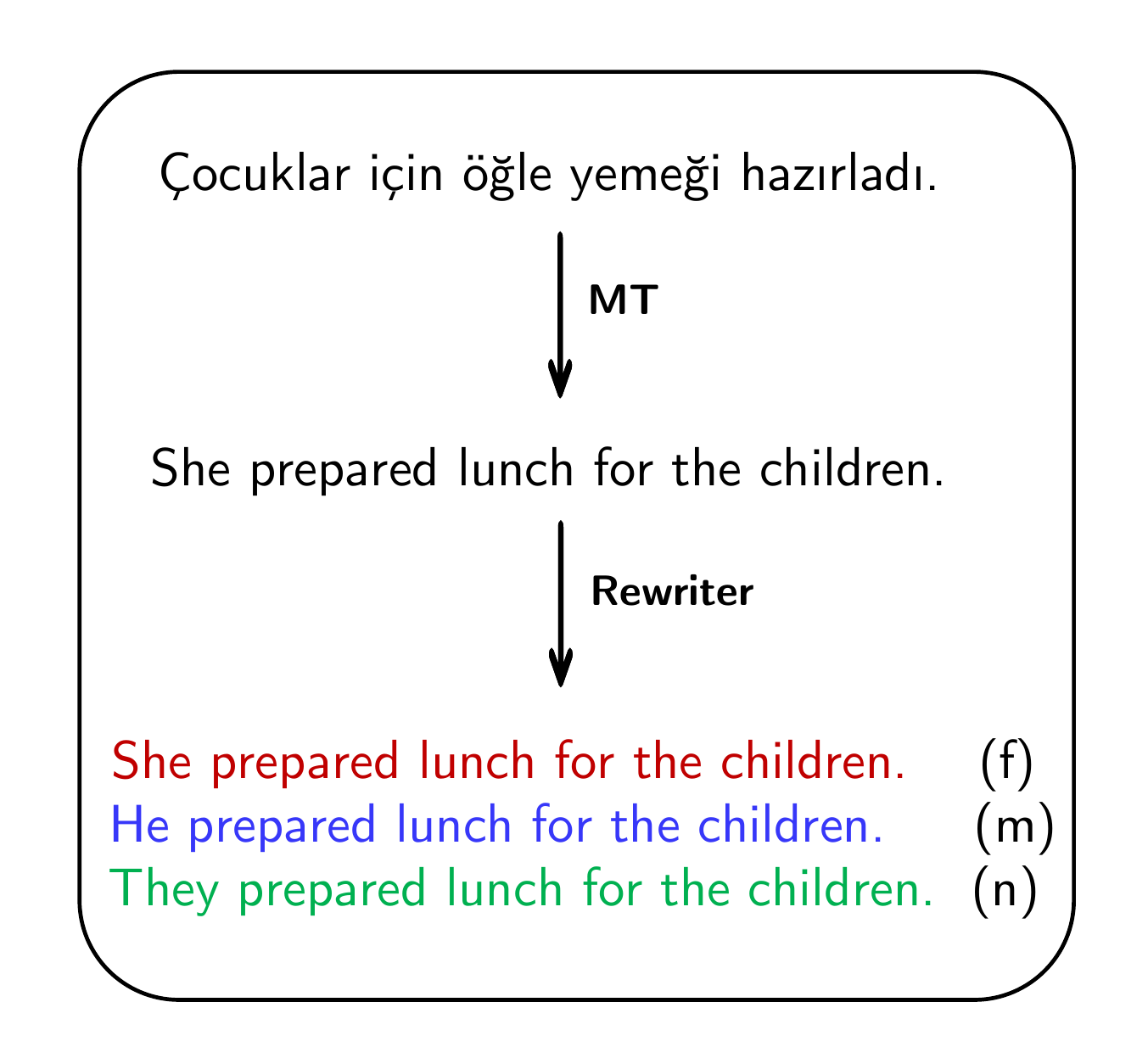}
\caption{ {\bf Gender Bias in Turkish-English Translation.} When translating from Turkish to English, the model tends to use the female pronoun \emph{she} for gender-unspecified individuals, likely due to a perceived link between women and child care. This bias can be mitigated by providing  feminine, masculine, and neutral rewrites.}
\label{fig:mt_bias_mitigation}
\end{figure}

This difference in gender on third-person singular pronouns leads to translation scenarios such as the one seen in Figure \ref{fig:mt_bias_mitigation}, where someone with no specified gender in the source is marked as female in the translation through the pronoun \emph{she}. NMT models often make gender assignments according to stereotypes \citep{stanovsky-etal-2019-evaluating} -  in this case a model appears to associate child care with women. One remedy for this category of problems is to supplement the default feminine translation with masculine and gender-neutral alternatives, so that all possible gender interpretations are covered. This can be accomplished by applying a gender rewriter to the original NMT output, as shown in the bottom portion of Figure \ref{fig:mt_bias_mitigation}.

GATE (Gender-Ambiguous Translation Examples, \citealt{rarrick2023gate}) introduced an evaluation benchmark for gender rewrites for translations from English into French, Spanish, and Italian. In this work, we introduce GATE X-E\footnote{X-E indicates translation from `X' language into English}, an extension to GATE that focuses on translations \emph{into} English from a set of more weakly-gendered languages. It consists of natural sentences with strong diversity of sentence lengths and domains, and challenges translation rewriters on a wide range of linguistic phenomena. GATE X-E contains between 1250 and 1850 instances for each of our language pairs.

We also present a translation-rewriting solution that utilizes GPT-4 \citep{chatgpt} to provide gendered and gender-neutral alternatives. It achieves high accuracy on the pronoun-only subset of GATE X-E, while. Finally, we also perform human evaluation and provide a detailed error analysis of the results.

The remainder of this paper is organized as follows -- In section \ref{sec:data_collection}, we discuss the corpus creation process and structure of GATE X-E. In section \ref{sec:rewriting_strat}, we discuss how various properties can affect the difficulty of translation rewriting problems. In section \ref{sec:Experiments} we introduce a GPT-4-based translation-rewriting solution and discuss how it is evaluated with GATE X-E. In section \ref{sec:Results} we discuss results of our experiments and perform detailed error analysis. Finally, In section \ref{sec:Related_Work} we cover related work.
\section{GATE X-E Dataset}
\label{sec:data_collection}

We introduce GATE X-E by describing the annotation process and labels used, as well as providing statistics on the collected data.

\subsection{Arbitrarily Gender-Marked Entities}

Following \citet{rarrick2023gate}, we use Arbitrarily Gender-Marked Entity (AGME) to refer to individuals whose gender is not marked in a source sentence, but is in a translation, either through a gendered pronoun or an intrinsically gendered noun. Presence of an AGME in a translation indicates that alternate gender translations are possible.

The subject pronoun from the example translation shown in Figure \ref{fig:mt_bias_mitigation} is an AGME. Because there is no gender marking in the source sentence, it is valid to translate the subject as \emph{she}, \emph{he} or \emph{they}.

\subsection{Dataset Creation and Annotation }

All instances in GATE X-E consist of a single source sentence with one or more translations covering possible gender interpretations. We pulled sentence pairs for each of our language pairs from several corpora found on OPUS\footnote{https://opus.nlpl.eu/}: Europarl \citep{koehn-2005-europarl}, TED talks \citep{TED-talks-corpus}, tatoeba\footnote{https://tatoeba.org/en/about}, wikimatrix \citep{schwenk-etal-2021-wikimatrix}, OpenSubtitles \citep{lison-tiedemann-2016-opensubtitles2016}, QED \citep{QED-corpus-citation} and CCAligned \citep{el-kishky-etal-2020-ccaligned}. We then apply the following filters:
\begin{itemize}
\item The source sentence scores at least 0.7 match for the intended language when using the python langdetect\footnote{https://pypi.org/project/langdetect/} package. 
\item The English translation contains at least one word on a curated word list consisting of 79 English nouns (e.g. \emph{mother, uncle, actress, duke}) and pronouns (\emph{he, she, him, her, his, hers, himself, and herself}.) This list is found in Table \ref{tab:gendered_noun_pronouns} in the appendix.
\end{itemize}

We then sampled sentences from the filtered set and provide them to annotators. From this data, the annotators selected appropriate sentences and annotated them for entity types, number of AGMEs, and gendered-alternative translations if AGMEs are present. To be included, a translation must include at least one gender-marked term in the target, which could be a pronoun\footnote{including possessive determiners \emph{his}, \emph{her}, \emph{their}} or noun\footnote{including other gender-marking modifiers, such as \emph{female}.}

For each language, a second annotator then reviewed the data to correct errors and inconsistencies. All of the annotators are native speakers of the source language, fluent in English, and hold advanced degrees in linguistics or a related field. 

If there are one or two AGMEs in the translated pair, they will provide translation variants so that all possible gender combinations for those AGMEs (among female, male, and neutral) are covered. They do so by replacing all gendered pronoun and noun mentions with corresponding words of the respective gender. Neutral variants are omitted if there is no suitable gender neutral term in common usage for a concept. For example, the term \emph{nibling} exists as a gender neutral variant of \emph{niece} or \emph{nephew}, but is not in common usage and so neutral variants of translations using \emph{niece} or \emph{nephew} would be left out. 

Some sentences may contain a mixture of references to AGMEs as well as to humans who are gender-marked in the source. In these cases, gender indicated in the source will be preserved in all translations, as in \emph{father} and \emph{his will} in the example shown in Figure \ref{mix_agmes}. In this example, \emph{Babası} explicitly indicates \emph{father} in the source.

\begin{table}[!h]
\tabcolsep 3pt
\centering
\begin{tabular}{lp{0.4\textwidth}}
\toprule
{\fontsize{10}{12}\bf Src}& {\fontsize{10}{12}Babası vasiyetinde arabayı ona bıraktı.} \\ 
{\fontsize{10}{12}\bf Fem}&{\fontsize{10}{12}Her father left her the car in his will.} \\  
{\fontsize{10}{12}\bf Masc}&{\fontsize{10}{12}His father left him the car in his will.} \\  
{\fontsize{10}{12}\bf Neut}&{\fontsize{10}{12}Their father left them the car in his will.}  \\ 
{\fontsize{10}{12}\bf Lbls}&{\fontsize{10}{12}target\_only\_gendered\_pronoun, source+target\_gendered\_noun+pronoun,
1-AGME, mixed}  \\ 
\bottomrule
\end{tabular}
\captionof{figure}{{\bf GATE X-E Example Instance.} This includes Turkish source; feminine, masculine and gender-neutral English translations; and labels.}
\label{mix_agmes}
\end{table}

\subsection{Labels}
The labels used in GATE X-E are defined in Table \ref{tab:label_definitions}, along with examples for each. All instances in GATE X-E refer to at least one person who is marked for gender in the English target. We include both positive and negative examples. In positive examples, at least one of those individuals was not marked for gender in the source, and is therefore an AGME, meaning that alternative translations with different gender markings are possible. In negative examples, all individuals who are marked for gender in the target are also marked in the source, so no alternative translations are possible.

\begin{table*}[p]
    \centering
    \begin{tabular}{p{2.85in}p{2.85in}}
    \toprule
    {\bf Description} & {\bf Example (tr $>$ en)} \\
    \midrule \midrule
    \multicolumn{2}{c}{\bf Negative/Non-AGME labels} \\
    \midrule
    \texttt{\bf source+target\_gendered\_noun} & \\
    A person is referred to by a gendered noun in both source and translations. & Git ve {\bf erkek kardeş}ine yardım et. $\rightarrow$ \newline Go and help your {\bf brother}. \\
    \midrule
    \texttt{\bf source+target\_gendered\_noun+pronoun} & \\ 
    A person referred to by a gendered noun in the source is referred to by both a gendered noun and one or more gendered pronouns in the translations. & {\bf Anne}m zaten karar{\bf ı}nı verdi. $\rightarrow$ \newline My {\bf mom} has already made {\bf her} decision. \\
    \midrule
    \texttt{\bf source\_gendered\_noun\_target\_pronoun} & \\ 
    A person is referred to by a gendered noun in the source, and one or more gendered pronouns in the translations (but not by a gendered noun). & {\bf O}, gerçek bir {\bf bilim adamı}dır. $\rightarrow$ \newline {\bf He} is a {\bf scholar} to the core. \newline (\emph{bilim adamı} indicates a male scholar) \\
    \midrule
    \texttt{\bf non-AGME-name} & \\
    A non-AGME person is referred to by name. & {\bf Umut}'un \emph{torunu} ünlü bir yazar değil mi?  $\rightarrow$ \newline {\bf Umut}'s \emph{granddaughter/grandson/grandchild} is a famous writer, isn't/aren't \emph{she/he/they}?\\
    
    \midrule \midrule
     \multicolumn{2}{c}{\bf Positive/AGME labels} \\
    \midrule
    \texttt{\bf target\_only\_gendered\_noun} & \\
    A person who is not gender-marked in the source is referred to with a gendered noun in the translations. & {\bf Yeğen}im bugün geliyor. $\rightarrow$ \newline My {\bf niece/nephew} is coming today. \\
    \midrule
    \texttt{\bf target\_only\_gendered\_pronoun} & \\
    A person who is not gender-marked in the source is referred to with a gendered pronoun in the translations. & {\bf Onun} yardımı paha biçilmezdi. $\rightarrow$ \newline {\bf Her/His/Their} help has been invaluable. \\
    \midrule
    \texttt{\bf target\_only\_gendered\_noun+pronoun} & \\
    A person who is not gender-marked in the source is referred to with both a gendered noun and gendered pronoun in the translations. & {\bf Torun}un iş{\bf ini} seviyor olmalı. $\rightarrow$ \newline Your {\bf granddaughter/grandson/grandchild} must love {\bf her/his/their} job. \\
    \midrule
    \texttt{\bf name} & \\
    An AGME is referred to by name. We treat personal names as non-gender-marking. & {\bf Beyza} akşam yemeğini bitiremedi. $\rightarrow$ \newline {\bf Beyza} wasn't able to finish {\bf her/his/their} dinner. \newline (\emph{Beyza} is typically considered a feminine name)\\
    \midrule \midrule
     \multicolumn{2}{c}{\bf Other} \\
    \midrule

    \texttt{\bf mixed} & \\
    Both positive and negative examples are present & \emph{Baba}{\bf sı} yine uçağ\emph{ını} kaçırdı. $\rightarrow$ \newline {\bf Her/His/Their} \emph{father} missed \emph {his} plane again.\\
    \midrule
    \texttt{\bf \emph{N} AGME(s)} & \\
    \emph{N} is a whole number representing the number of AGMEs in the instance. Negative examples are annotated as 0 AGMEs.& 
    0 AGME: My mother read her book. \newline
    1 AGME: {\bf She/He} ate {\bf her/his} lunch alone. \newline
    2 AGME: {\bf She/He} annoyed \emph{her/him} with 
    {\bf her/his} music.
    \\
    \bottomrule
    \end{tabular}
    \caption{{\bf Label Definitions and Examples.} Words relevant to the label are bolded or italicized in source and target. \emph{Pronoun} in these definitions includes possessive determiners \emph{her}, \emph{his}, \emph{their}.}
    \label{tab:label_definitions}
\end{table*}

\subsection{Corpus Statistics}

Table \ref{tab:label_counts} in the appendix provides a comprehensive breakdown of corpus statistics for GATE X-E, with instance counts per language for each label.

More than half of the instances for each language pair have exactly one AGME, with around 20-30\% are negative instances, having no AGMEs at all. Most AGMEs are \texttt{target\_only\_gendered\_pronoun}, meaning that they have no gender markings in the source and the only words in the target that mark their gender are pronouns. This is in part because there are relatively few nouns which are gendered in English but not gendered in the source languages.

Non-AGME references will involve a gendered noun on the source, and for most of the languages about half of these also include a pronoun reference.

Each language pair contains around 250 instances labeled \texttt{mixed}. These instances contain at least one AGME and at least one individual who is marked for gender in the source sentence. \texttt{name} indicates that there is an AGME who is referred to by name. Between 15 and 25\% of instances per language have this label. texttt{non-AGME-name} indicates that a non-AGME person is referred to by name. We present the distribution of sentence lengths in source and target languages in Figure \ref{fig:mt_bias_mitigation_2}.
\section {Translation Gender Rewriting}
\label{sec:rewriting_strat}

Translation gender rewriting is the process of taking a translated source-target pair and producing alternative translations with different gender markings. In a correctly rewritten translation, the gender markings should remain compatible with all gender information found in the source sentence \citep{habash-etal-2019-automatic}.
We consider this problem from the viewpoint of a user who wishes to see a set of three gendered-alternative translations with uniform output gender side-by-side: all-female, all-male and all-neutral. Because the translations will be viewed as a set, the translations should only vary from one-another in specific words that mark gender.

Here we discuss the difference in difficulty between rewrite problems where gendered nouns are included and those where gender is only marked by pronouns.

\vspace{.25cm}
\subsubsection*{Pronoun-Only Problems}
\label{sec:pronoun-only-problems}
For our source languages, if the only gender markers in the target sentence are gendered pronouns, there typically cannot be gender markers in the source sentence, since those languages do not have any gendered pronouns. We can therefore assume that if we have no gender information from external context, then all individuals mentioned by a gendered pronoun in the translation must be AGMEs. 

Rewriting in this scenario reduces to the relatively simple task of adjusting surface forms of all pronouns to match the desired gender. For rewrites involving \emph{he} or \emph{she} to gender-neutral \emph{they}, some verbs must additionally be adjusted to compatible surface forms. Since our focus is on rewrites with uniform gender assignments in the output (all-female, all-male, or all-neutral), this removes any need to determine which pronouns refer to which individual where more than one is mentioned. 

Here we see an example of a Pronoun-only instance with two AGMEs:

\begin{table}[!h]
\tabcolsep 3pt
\centering
\begin{tabular}{ll}
\toprule
{\fontsize{10}{12}\bf Female + Female}& {\fontsize{10}{12}She gave her her umbrella.} \\ 
{\fontsize{10}{12}\bf Male + Male}&{\fontsize{10}{12}He gave him his umbrella.} \\  
{\fontsize{10}{12}\bf Neutral + Neutral}&{\fontsize{10}{12}They gave them their umbrella.}\\
\bottomrule
\end{tabular}
\end{table}

Instances of GATE X-E that fall into this subtype will always have the label \texttt{target\_gendered\_pronoun\_only}, and never any labels with containing \texttt{gendered\_noun}.

Note that there is an exceptional scenario, where external context makes it reasonable for a gendered noun in one of our source languages to be translated into a gendered pronoun in English. See the example for \texttt{source\_gendered\_noun\_target\_pronoun} in Table \ref{tab:label_definitions}, and refer to Appendix \ref{sec:appendix} for further discussion.

\vspace{.25cm}
\subsubsection*{Gendered-Noun Problems} If we expand our scope to include translations containing gendered nouns, we encounter several new challenges that render the rewriting problem significantly more difficult. The following pair of examples illustates some of those new challenges.

\vspace{.25cm}
\begin{center}
\emph{Kardeşine ziyarete gelip gelmeyeceğini sordu.}

$\big\Downarrow$ 

\emph{He asked his sister if she would visit.}
\end{center}

In this translation, both the male and female individuals are AGMEs since \emph{Kardeşine} simply denotes a \emph{sibling} without any gender specification. Therefore, there are nine possible rewrites, including the original translation: \emph{He} and \emph{his} can be optionally replaced with \emph{she/her} or \emph{they/their}, and \emph{sister/she} can be optionally replaced with \emph{brother/he} or \emph{sibling/they}.

In the next example, however, the gender of the sibling is specified in the source as female by the addition of the word \emph{Kız}, even though the default English translation is exactly the same:
\vspace{.25cm}
\begin{center}
\emph{\textbf{Kız} kardeşine ziyarete gelip gelmeyeceğini sordu.}

$\big\Downarrow$ 

\emph{He asked his sister if she would visit.}
\end{center}
\vspace{.25cm}

Here \emph{sister} must remain fixed because of the gender marking in the source. \emph{She} is also fixed because it is coreferent with the sister. Only the individual referred to by \emph{he} and \emph{his} is an AGME, so only three valid rewrites exist (including the original):

\begin{table}[!h]
\tabcolsep 3pt
\centering
\begin{tabular}{ll}
\toprule
{\fontsize{10}{12}\bf Female}&{\fontsize{10}{12}She asked her sister if she would visit.} \\  
{\fontsize{10}{12}\bf Male}& {\fontsize{10}{12}He asked his sister if she would visit.} \\ 
{\fontsize{10}{12}\bf Neutral}&{\fontsize{10}{12}They asked their sister if she would visit.}\\
\bottomrule
\end{tabular}
\end{table}

More specifically, the additional challenges inherent in this problem class include the following. 

\begin{itemize}

\item Gendered-marked nouns may appear on the source as well, so we must examine both source and target to determine if variants are needed. This is demonstrated by the behavior of \emph{sister} in the two examples above.
\item Gendered pronouns in the target may refer to individuals whose gender was marked in the source, and are therefore not appropriate to modify. When multiple individuals are mentioned, we must differentiate which ones refer to such non-AGME individuals to produce a correct rewrite. This is demonstrated by the behavior of \emph{she} in the two examples above.

\end{itemize}

A system that is capable of solving these problems must then able to implicitly perform coreference resolution and alignment of nouns between the source and target, significantly increasing the complexity of a solution.

\section{Experiments}
\label{sec:Experiments}

We introduce a novel translation gender-rewriting solution that leverages GPT-4 and evaluate it on GATE X-E. 

\subsection{Rewriting with GPT-4}

Our solution uses chain-of-thought prompting \citep{chain-of-thought-wang-etal} to elicit GPT-4 to produce three variant translations for each input source-translation pair -- all-neutral, all-female and all-male, while leaving any gendered words associated with non-AGMEs unmodified. 

Each step in the prompt is accompanied by detailed clarifications and example vocabulary. The prompt also includes three full examples, customized per source language. The examples indicate that "None" should be returned in lieu of translation variants when there are no AGMEs present. The full prompt for Turkish-English can be found in the appendix in Figures \ref{fig:few-shot-gpt-4-part-1} and \ref{fig:few-shot-gpt-4-part-2}.

We process extract the three variants from GPT's output. The output in some cases is "None", in which case we treat all outputs as identical to the original translation.

\subsection {Data Preparation}
\label{sec:data_preparation}

Each GATE X-E instance consists of a source sentence and a set of translations in English. We pull a set of sentence-tuples from each instance for use in evaluation as test tuples. Test tuples consists of a source sentence, an original translation, and a reference for a specific set of gender assignments. We group these by the gender-assignments of AGMEs in the original translation, producing four subsets: feminine, masculine, mixed feminine and masculine (with 2 AGME), and negative.

For 1-AGME instances, we extract four tuples: feminine(f)${\rightarrow}$masculine(m), f${\rightarrow}$neutral(n), m${\rightarrow}$f, and m${\rightarrow}$n. For 2-AGME instances, we additionally extract 6 tuples covering both combinations of mixed-gender original translations and all three sets of uniform gender assignments in the reference (\{f+m, m+f\}$\rightarrow$\{f+f, m+m, n+n\}).

Negative instances, which do not include AGMEs and should not be modified by the rewriter, are handled separately. For each negative instance we create tuples to check that the original target remains unmodified in masculine, feminine and neutral outputs. 

For simplicity, we exclude the handful of instances with three or more AGMEs. We do not include any tuples that include gender-neutral forms in the original target because because of the ambiguity in distinguishing singular neutral pronouns them plural \emph{they} forms. For any instance where the neutral reference is empty (e.g. because there is no neutral form of term), no test tuple with neutral reference is created. 

As we are interested in exploring the difference in behavior of our solution we also mark each test tuple according to whether they contain gendered nouns in either source or target, or only gendered pronouns in the target. The former group we designate as \emph{Gendered-Noun} and the latter group \emph{Pronoun-Only}. Note that negative examples always contain a gendered noun in the source, and so they do not have a distinction between the two types.

Table \ref{tab:complexity_counts} in the appendix shows counts for each source language and category. Note that tuples with uniform gender original translations (f/m) are always created in pairs, so they have the same counts.

\section{Results}
\label{sec:Results}

\begin{table*}
\centering
\small
\begin{tabular}{lllcccccccc}
\toprule
    \multirow{2}{*}{\bf Language} & \multirow{2}{*}{\bf Subtype } & \multicolumn{2}{c}{{\bf Fem Orig $\uparrow$}} & \multicolumn{2}{c}{{\bf Masc Orig $\uparrow$}}  & \multicolumn{3}{c}{{\bf  Mixed Orig  $\uparrow$}} & \multicolumn{2}{c}{{\bf  Negative  $\uparrow$}}\\
\cmidrule(lr){3-4} \cmidrule(lr){5-6} \cmidrule(lr){7-9} \cmidrule(lr){10-11}
    & & {\bf M} & {\bf Neut} & {\bf F } & {\bf Neut} & {\bf F} & {\bf M} & {\bf Neut} & {\bf Gender} & {\bf Neut}\\
\midrule
\midrule
   \multirow{3}{*}{tu $\rightarrow$ en} 
     & Overall & 0.81 & 0.86 & 0.80 & 0.85 & 0.78 & 0.82 & 0.57 & 0.87 & 0.80 \\
     & Pronoun-Only  & 0.99 & 0.99 & 0.96 & 0.99 & 0.93 & 0.96 & 0.98 & - & - \\
     & Gendered-Noun & 0.59 & 0.63 & 0.60 & 0.63 & 0.75 & 0.80 & 0.45 & - & - \\    
    \hline
    \multirow{3}{*}{fi $\rightarrow$ en} 
     & Overall & 0.90 & 0.80 & 0.82 & 0.81 & 0.67 & 0.83 & 0.63 & 0.89 & 0.88 \\
     & Pronoun-Only   & 0.98 & 0.98 & 0.97 & 0.98 & 0.94 & 0.96 & 0.97 & - & - \\
     & Gendered-Noun & 0.75 & 0.53 & 0.54 & 0.57 & 0.34 & 0.68 & 0.54 & - & - \\  
\hline
    \multirow{3}{*}{hu $\rightarrow$ en} 
     & Overall & 0.85 & 0.79 & 0.84 & 0.81 & 0.75 & 0.92 & 0.78 & 0.84 & 0.84 \\
     & Pronoun-Only   & 0.98 & 0.99 & 0.96 & 0.98 & 0.92 & 0.97 & 0.97 & - & - \\
     & Gendered-Noun & 0.66 & 0.61 & 0.68 & 0.76 & 0.56 & 0.86 & 0.56 & - & - \\
\hline
    \multirow{3}{*}{fa $\rightarrow$ en} 
     & Overall & 0.86 & 0.81 & 0.86 & 0.81 & 0.81 & 0.84 & 0.66 & 0.54 & 0.53 \\
     & Pronoun-Only   & 0.99 & 0.98 & 0.99 & 0.98 & 0.93 & 0.93 & 0.96 & - & - \\
     & Gendered-Noun & 0.67 & 0.54 & 0.68 & 0.55 & 0.78 & 0.81 & 0.58 & - & - \\
\bottomrule
\end{tabular}
\caption{{\bf Accuracy of our Rewriting Solution.} Accuracy on test elements for each source language, problem subtype, original target gender (top header row), and requested output gender (second header row). Only exact matches to reference are counted.}
\label{tab:gpt_41_results}
\end{table*}

Table \ref{tab:gpt_41_results} shows the accuracy of our solution on test elements over each combination of source language, and subtype level, with \emph{Overall} indicating an aggregate score over \emph{Pronoun-Only} and \emph{Gendered-Noun} elements. The top labels in the header row indicate gender of AGMEs in the original target, and the bottom indicates the desired output gender. 

Following \citet{rarrick2023gate}, we focus on exact match accuracy to the reference. Frequently only one or two words will be different between an original translation and a correct rewrite. In this context, metrics such as BLEU \citep{papineni2002bleu} and WER are not very effective at determining the significance of single extraneous or missed word modification.

\subsection{Pronoun-Only Subset}

On the Pronoun-Only subset, the solution rarely makes mistakes for masculine and feminine original targets, with scores ranging from 0.96 to 0.99. Test cases where the original target has mixed gender all come from 2-AGME instances. These skew towards longer and more complicated sentence, which thus leads to slightly lower accuracy.

On most language pairs we see that Pronoun-Only rewrites into masculine outperform rewrites into feminine by a few percentage points. The largest gap is 5 points for mixed-gender original target on Hungarian. This may indicate a slight general tendency of GPT-4 to prefer phrasing using masculine pronouns.

\subsection{Gendered-Noun Subset}

Scores on the Gendered-Noun subset are substantially lower than for \emph{Pronoun-Only}, generally ranging from about 0.5 to 0.8, with Finnish mixed$\rightarrow$feminine as an outlier at the low end at 0.34. The score differential with \emph{Pronoun-Only} can be mostly attributed to the more difficult nature of the problems. However, there are some cases where there are multiple acceptable alternative phrases to use in a rewrite, and GPT-4 chooses a different one from the reference. 

This effect is most pronounced in the Finnish \emph{Complex} mixed-original target data. This subset contains a large amount of data from Europarl that includes titles such as \emph{Mr.} and \emph{Mrs.} and addresses to \emph{Mr. President}. The feminine rewrites often choose a mismatched form, such as \emph{Ms. Müller} rather than \emph{Mrs. Müller}, or \emph{Mrs. President} rather than \emph{Madam President}. 

Similarly, neutral rewrites on such sentences often produces \emph{Honorable President}, \emph{Chairperson}, or \emph{Honorable Speaker} as a rewrite of \emph{Mr. President} or \emph{Madam President}, mismatching \emph{Honored President} in the reference. 

\subsection{Negative Subset}

\begin{table*}
\centering
\begin{tabular}{lcccccc}
\toprule
 \multirow{2}{*}{\textbf{Error Type}} 
 & \multicolumn{2}{c}{{\bf  Pronoun-Only  $\uparrow$}} 
 & \multicolumn{2}{c}{{\bf  Gendered-Noun  $\uparrow$}} 
 & \multicolumn{2}{c}{{\bf  Negative  $\uparrow$}} \\
 
\cmidrule(lr){2-3} \cmidrule(lr){4-5} \cmidrule(lr){6-7}
   & \multicolumn{1}{c}{\bf Gen (\%)} & \multicolumn{1}{c}{\bf Neut (\%)} & \multicolumn{1}{c}{\bf Gen (\%)} & \multicolumn{1}{c}{\bf Neut (\%)} & \multicolumn{1}{c}{\bf Gen (\%)} & \multicolumn{1}{c}{\bf Neut (\%)}\\
\midrule
\midrule
extraneous noun change      & 12.5       & 10.0        & 2.8       & 3.5      & 0.0          & 34.5     \\
extraneous pronoun change  & 0.0           & 10.0        & 4.0        & 5.6      & 100           & 65.5      \\
missing noun change    & 0.0           & 0.0         & 55.5       & 51.9      & \multicolumn{1}{c}{-}           & \multicolumn{1}{c}{-}          \\
missing pronoun change & 87.5       & 80.0       & 37.5       & 38.6      & \multicolumn{1}{c}{-}           & \multicolumn{1}{c}{-}  \\
\midrule
total errors  & \multicolumn{1}{c}{30}          & \multicolumn{1}{c}{8}          & \multicolumn{1}{c}{427}         & \multicolumn{1}{c}{381}        & \multicolumn{1}{c}{57}          & \multicolumn{1}{c}{58} \\
\bottomrule
\end{tabular}
\caption{\textbf{Distribution of Errors from Human Evaluation for Turkish-English.} Shows percentages of errors over Pronoun-Only, Gendered-Noun and Negative subsets of the data, for gendered and neutral requested outputs.}
\label{tab:human_annotation}
\end{table*}

\emph{Negative$\rightarrow$Gender} score indicates how often both the female and male variants produced by our solution exactly matched the original translation, while \emph{Negative$\rightarrow$Neutral} measures the same for the neutral variant. An output of \emph{None} from GPT-4 is also possible, indicating that all variants should be considered a copy of the original translation. On the negative data subset, this is considered a reference match.

For Turkish, Finnish and Hungarian, we see scores for both \emph{Gender} and \emph{Neutral} subsets in the 0.8 to 0.9 range. Farsi is an outlier with 0.54 and 0.53 at the low end.

With the exception of Turkish, we find that we almost never see matches on negative test items aside from \emph{None} outputs.  For Turkish, 47\% of non-\emph{None} outputs match for Gendered and 15\% of neutral outputs. For all languages, however, we do see a large number of neutral outputs that match a version of the original translation where all pronouns are modified to their neutral variants\footnote{and verb agreement modified to match}. This occurs in particular when there are gendered nouns in the source which determine the gender of some pronouns in the translation as well. For example, instead of \emph{the man has something under his coat}, it would output \emph{the man has something under \textbf{their} coat}. 

If were relax matching criteria to allow this variant, neutral negative accuracy increases to 0.91 for Farsi, 0.92 for Turkish, 0.95 for Finnish and 0.95 for Hungarian.

\subsection{Human Evaluation}

For the Turkish data, one of the Turkish annotators provided annotations for error type on all outputs that did not exactly match the reference. These are found in Table \ref{tab:human_annotation}. 

For each test item, they mark whether there were nouns or pronouns were changed where the reference and original translation matched (\emph{extraneous noun/pronoun change}, as well as whether any nouns or pronouns should have been changed but were not (\emph{missing noun/pronoun change}). Distributions for masculine and feminine outputs were similar, so we show combined \emph{gendered} outputs for each subtype. We also aggregate over mixed- and uniform-gender inputs. 

For positive test cases, missing noun and pronoun changes were far more common than extraneous changes. For cases containing gendered nouns, noun changes were missed more often than pronoun changes, while extraneous pronoun changes were more common than extraneous noun changes.

For Negative test cases, gendered output only ever contained extraneous pronouns changes, while neutral outputs did have a fair number of extraneous noun changes. An example of these is changing \emph{man} to \emph{person} even when \emph{man} was gender-marked in the source sentence.

Among missing pronoun errors, missing possessive determiners was by for the most, with subject and object pronoun errors roughly equivalent. Missed and extraneous reflexives were extremely rare. We also saw a single of subject-verb agreement error each when changing \emph{he} and \emph{she} to \emph{they}, and one case where other wording was changed in the sentence.

\section{Related Work}
\label{sec:Related_Work}
\noindent\textbf{Understanding and Assessing Gender Bias}: It has been documented that machine translation engines often make mistakes and show gender biases when translating between languages with differing gender norms \citep{stanovsky-etal-2019-evaluating,prates2019assessing,rescigno2020case, lopez2021gender,prates2019assessing,fitria2021}.

\noindent\textbf{Evaluation Benchmarks}: 
\textit{Translating from English}: \citet{bentivogli-etal-2020-gender} and \citet{savoldi-etal-2022-morphosyntactic} introduced the MuST-SHE dataset, which includes triplets of audio, transcript, and reference translations for English to Spanish, French, and Italian languages, classified by gender. \citet{stanovsky-etal-2019-evaluating} developed the WinoMT challenge set, which includes English sentences with two animate nouns, one of which is coreferent with a gendered pronoun. 

\citet{renduchintala-etal-2021-gender} introduced the SimpleGEN dataset for English-Spanish and English-German language pairs, which includes short sentences with occupation nouns and clear gender indications. The Translated Wikipedia Biographies dataset includes human translations of Wikipedia biographies for gender disambiguation evaluation. Lastly, 
\citet{currey-etal-2022-mt} presented the MT-GenEval dataset, which includes gender-balanced, counterfactual data in eight language pairs, specifically focusing on translation from English into eight widely-spoken languages.

\textit{Translating into English}: Numerous studies have focused evaluating bias in translating from from a weakly gendered language such as Turkish into English. \citep{prates2019assessing,fitria2021,ciora2021examining,ghosh2023chatgpt}. 

\noindent\textbf{Strategies for Gender Bias Mitigation }: To mitigate gender bias when translating queries that are gender-neutral in the source language, Google Translate announced a feature \citep{kuczmarski2018,johnson2020} that provides gender-specific translations. Both \citet{sun2021they} and \citet{vanmassenhove-etal-2021-neutral} have explored monolingual gender-neutral rewriting of English demonstrating that a neural model can perform this task with reasonable accuracy. \citet{ghosh2023chatgpt} evaluates gender bias in GPT-3.5 Turbo output when translating from gender-neutral languages into English. To the best of our knowledge, our work is the first to leverage GPT-4 for mitigation.
\section{Conclusion}

We have presented GATE X-E, a diverse dataset covering a wide range of scenarios relevant to translation gender-rewriting for English-target language pairs, covering gendered and gender-neutral rewrites. We have discussed intricacies of the English-target translation rewrite problems, and explained what properties lead to easier or more difficult rewrite problems. We have explored the ability of GPT-4 to provide rewrites for these translations, showing that it can achieve very high accuracy on a pronoun-only rewriting problems, but performs less well when gendered nouns are introduced.

We also hope that by making GATE X-E accessible to the broader research community, we can encourage further research on gender debiasing in the machine translation space.

\section{Limitations}
Our study has some limitations that could be addressed in future research. Firstly, while we utilized GPT-4 for rewriting tasks, the potential of open-source models remains unexplored and could be beneficial. Secondly, our rewriter operates on few-shot chain-of-thought prompting. Future investigations could consider exploring the zero-shot setting, which could potentially be more cost-efficient.

\bibliography{anthology,custom}

\appendix
\label{sec:appendix}
\section{Further details on GATE X-E}
\label{sec:further_details}

\begin{table*}
    \centering
    \begin{tabular}{lrrrr}
    \toprule
        & \bf{tr $\rightarrow$ en} & \bf{fa $\rightarrow$ en} & \bf{fi $\rightarrow$ en} & \bf{hu $\rightarrow$ en} \\
        \midrule
        \midrule
        \bf{total instance count} & 1,429 & 1,259 & 1,832 & 1,308 \\
        \midrule
        target\_only\_gendered\_noun & 142 & 118 & 159 & 95 \\ 
        target\_only\_gendered\_pronoun & 1,074 & 906 & 1,096 & 914 \\
        target\_only\_gendered\_noun+pronoun & 114 & 49 & 105 & 115 \\
        source+target\_gendered\_noun & 239 & 244 & 379 & 75 \\ 
        source+target\_gendered\_noun+pronoun & 328 & 292 & 361 & 422 \\ 
        source\_gendered\_pronoun\_target\_noun & 3 & 0 & 0 & 33 \\ 
        \midrule
        0 AGMEs & 300 & 264 & 502 & 264 \\
        1 AGME & 900 & 869 & 1,164 & 848 \\ 
        2 AGMEs & 225 & 124 & 161 & 192 \\ 
        3 AGMEs & 4 & 2 & 5 & 4 \\
        \midrule
        mixed & 271 & 263 & 237 & 262 \\
        name & 328 & 175 & 408 & 159 \\
        non-AGME-name & 32 & 5 & 136 & 16 \\
        \bottomrule
    \end{tabular}
    \caption{{\bf GATE X-E Statistics.} Sentence counts per language associated with each label.}
    \label{tab:label_counts}
\end{table*}

Table \ref{tab:label_counts} provides a comprehensive breakdown of corpus statistics for GATE X-E, with instance counts per language for each label.

\begin{figure*}
\centering \includegraphics[width=0.9\linewidth]{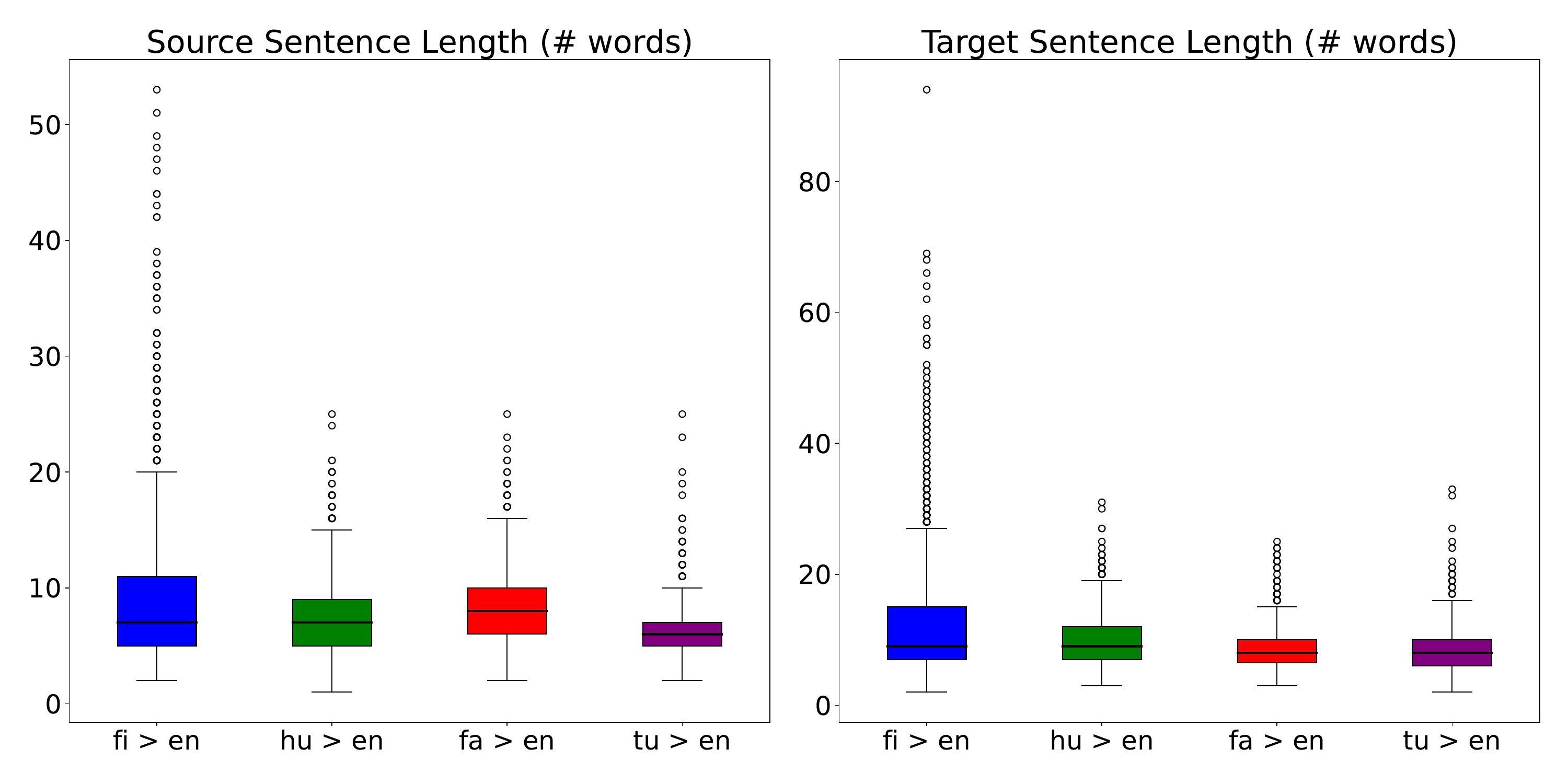}
\caption{ {\bf Boxplots representing the distribution of sentence lengths in source and target languages.} The four language pairs are Finnish to English (fi > en), Hungarian to English (hu > en), Persian to English (fa > en), and Turkish to English (tu > en). The left plot represents the source language sentence lengths, and the right plot represents the target language (English) sentence lengths. The color of each boxplot corresponds to the language pair as indicated in the legend.}
\label{fig:mt_bias_mitigation_2}
\end{figure*}

Figure \ref{fig:mt_bias_mitigation_2} presents boxplots that demonstrate the sentence length distribution on source and target for four language pairs: Finnish to English, Hungarian to English, Persian to English, and Turkish to English. The left plot shows the sentence lengths in the source languages, and the right plot displays the sentence lengths in English, the target language. The legend indicates the color corresponding to each language pair. Compared to the other three language pairs, Finnish to English contains longer sentences.

Table \ref{tab:complexity_counts} shows counts for each source language and category. Note that tuples with uniform gender original translations (f/m) are always created in pairs, so they have the same counts.

\begin{table}
    \centering
    \begin{tabular}{lrrrr}
    \toprule
        \bf{Category} & \bf{tr} & \bf{fa} & \bf{fi} & \bf{hu} \\
        \midrule
        \midrule
        Pronoun-Only (f/m) & 628 & 857 & 590 & 580 \\ \hline
        Gendered-Noun (f/m) & 500 & 473 & 454 & 415 \\ \hline
        Pronoun-Only (mix) & 54 & 180 & 198 & 44 \\ \hline
        Gendered-Noun (mix) & 392 & 142 & 186 & 200 \\ \hline
        Negative & 300 & 502 & 264 & 264 \\
        \bottomrule
    \end{tabular}
    \caption{{\bf Test tuple Counts By Category and Source Language.} Counts of pronoun-only, gendered-noun and  negative test tuples per source language. f/m signifies count for uniform gender original targets, while mix signifies a mixture of male and female references in the original target.}
    \label{tab:complexity_counts}
\end{table}

\section{Monolingual Rewriting with GPT-3.5 Turbo}
\label{sec:monolingual_rewriting}
GPT-4 performs very well on the pronoun-only subset of examples. However, its inference cost is high. Therefore, we evaluate the pronoun-only subset using GPT-3.5 Turbo.

\subsection{Gender-Neutral Rewriting}

As shown in \ref{sec:pronoun-only-problems}, pronoun-only uniform-gender do not require access to source information, and so our GPT-3.5 Turbo solution is only given access to original translation target. We first use GPT-3.5 Turbo to produce an all-neutral rewrite and then use a rule based solution to convert the all-neutral rewrite to gendered rewrites. 

The trickiest aspects of the gender-neutral rewrite are disambiguating pronoun classes for \emph{her} and \emph{him}, and adjusting verb forms when subjects change from \emph{she/he} to \emph{they}, so these are the primary decisions that GPT-3.5 Turbo must make.
 
We experiment with zero-shot and few-shot approaches. The zero-shot approach uses a single-sentence prompt as seen in Figure \ref{fig:zero-shot}. The few-shot approach expands on this prompt by adding five examples, and it can be seen in Figure \ref{fig:few-shot-turbo}.

\subsection{Gender-Neutral to Gendered Rewriting}

A useful simplifying observation is for uniform-gender pronoun only rewrites, we can generate a correct feminine or masculine rewrite from the original target and a correct all-neutral rewrite. Referring back to \ref{tab:pronoun_cats}, we see that all elements in the neutral column are unique, while the masculine and feminine columns each have one surface form fitting two categories. Knowing the correct neutral pronoun  fully determines what pronouns should be used for a given gender.

In practice the neutral rewrite may contain errors. To minimize their impact on the gendered rewrites, we begin with the original translation, and map to pronouns directly to the desired gender where unambiguous. When ambiguous (i.e. for \emph{her} or \emph{his} in the original target), we rely on the chosen form of the neutral pronoun to disambiguate.

\begin{figure}[t]
\centering \includegraphics[width=0.9\linewidth]{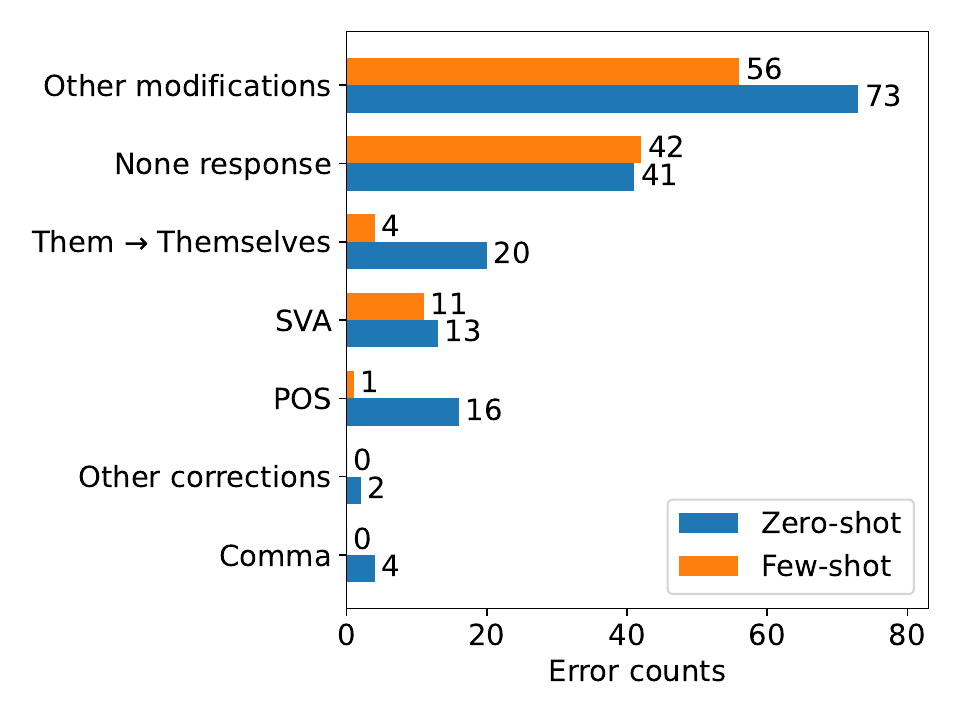}
\caption{\textbf{Distribution of errors in GPT-3.5 Turbo's zero-shot and few-shot settings.} The majority of errors in both settings stem from unrelated modifications and the model's 'None' response, indicating no need for gender-neutral rewriting.}
\label{fig:error_bar_plot}
\end{figure}

\begin{table*}
\centering
\small
\begin{tabular}{llcccccccc}
\toprule
    {\bf Language Pair} & {\bf Method} & \multicolumn{3}{c}{\bf Neutral Rewriting} & \multicolumn{3}{c}{\bf Gendered Rewriting} \\
    \cmidrule(lr){3-5} \cmidrule(lr){6-8}
    & & {\bf Accuracy (\%) $\uparrow$} & {\bf BLEU $\uparrow$ } & {\bf WER $\downarrow$}  & {\bf Accuracy (\%) $\uparrow$} & {\bf BLEU $\uparrow$ } & {\bf WER $\downarrow$}\\
\midrule
\midrule
    \multirow{3}{*}{tr $\rightarrow$ en } &  \citeauthor{sun2021they} \citeyear{sun2021they}&  96.16 & {\bf 99.65 } & 0.53 & - & - & - \\
    & Zero-shot & 97.24 & 99.30  &  0.80 & {\bf 99.50 } & {\bf 99.90 } & {\bf 0.90} \\
    & Few-shot & {\bf 98.90} &  99.55 & {\bf 0.44} &   99.46 &  99.00 &  0.10\\
    \hline
    \multirow{3}{*}{hu $\rightarrow$ en} & \citeauthor{sun2021they} \citeyear{sun2021they}  & 96.14  & {\bf 99.66 } & {\bf 0.53 } & - & - & - \\
     & Zero-shot & 96.58 &  99.04 &  1.27 & {\bf 99.27 }&  {\bf 99.95} &  {\bf 0.08}\\
    & Few-shot & {\bf 97.00} &  99.03 &  1.20 &  99.20 & 99.94 &  0.09 \\
        \hline
    \multirow{3}{*}{fi $\rightarrow$ en} & \citeauthor{sun2021they} \citeyear{sun2021they}  & 95.24 & {\bf 99.63 } &  {\bf 0.62 } & - & - & -\\
     & Zero-shot & 94.80 &  98.61 &  1.75 & 98.41 &  {\bf 99.85 } &  0.24 \\
    & Few-shot & {\bf 96.77 } &   98.62 &  1.54 & {\bf 98.99 } &  99.80 & {\bf 0.19} \\
        \hline
    \multirow{3}{*}{fa $\rightarrow$ en} & \citeauthor{sun2021they} \citeyear{sun2021they}  &  94.43 &  {\bf 99.57 } & {\bf 0.65 } & - & - & - \\
     & Zero-shot &  95.59 &  99.00  &  1.11 &  98.75 &  99.91 &  1.13 \\
    & Few-shot & {\bf 97.84} &  99.16 &  1.01 & {\bf 99.00} & {\bf 99.93} & {\bf 0.09}\\
\bottomrule
\end{tabular}
\caption{{\bf Results of Gender Neutral and Gendered Rewriting  on the Pronoun-Only Subset of GATE X-E}. We report the performance of the rule-based system proposed by \citeauthor{sun2021they} \citeyear{sun2021they}. Additionally, we evaluate GPT-3.5 Turbo in both zero-shot and few-shot settings. Gendered alternatives are generated using the algorithm described in Section \ref{sec:monolingual_rewriting}}
\label{tab:gender_neutral_results}
\end{table*}

\subsection{Experiments}

\subsubsection {  Rewriters }
\noindent{\bf Neutral rewriter Systems}:
 We consider the following rewriting systems:
\begin{enumerate}[leftmargin=*,itemsep=0.0ex, parsep=0pt] 
\item Rule-based system proposed by \citet{sun2021they}: It uses Spacy and GPT-2 to resolve ambiguity with \emph{his} and \emph{her}, and to adjust verb forms as needed. They also trained a neural model, but it was unfortunately not accessible.
\item GPT-3.5 Turbo: We evaluate GPT-3.5 Turbo on zero-shot and few-shot settings, using the prompts shown in Figures \ref{fig:zero-shot} and   \ref{fig:few-shot-turbo} in the appendix. 
\end{enumerate}
We investigated the neural model introduced by \citet{vanmassenhove-etal-2021-neutral}  as well, but were unable to reproduce results on their test data.

For all GPT-based rewrites we set temperature $T=0$
\subsubsection {  Evaluation }
We report the rewriter systems' performance using BLEU \citep{papineni2002bleu}, Word Error Rate (WER), and Accuracy.

In the gender-neutral rewriting task (Table \ref{tab:gender_neutral_results}), GPT-3.5 Turbo performs better in the few-shot setting compared to the zero-shot setting. Although GPT-3.5 Turbo provides slightly higher accuracy compared to the rule-based system proposed by \citet{sun2021they}, the rule-based system performs better based on BLEU and WER. This is because GPT-3.5 Turbo makes modifications unrelated to neutral rewriting, as detailed in the error analysis section.

In the gendered-alternatives rewriting task (Table \ref{tab:gender_neutral_results}), the zero-shot setting indicates that for resolving the \emph{her}$\rightarrow$\emph{his/him} and \emph{his}$\rightarrow$\emph{her/hers} ambiguity, gender-neutral rewrites from the zero-shot prompt are used. Similarly, the few-shot setting uses the corresponding gender-neutral outputs from the few-shot prompt. The performance of both settings is comparable.
\subsubsection{Error Analysis}

Figure \ref{fig:error_bar_plot} illustrates the distribution of aggregated errors across four language pairs for GPT-3.5 Turbo in both zero-shot and few-shot settings, specifically for the task of gender-neutral rewriting. The definitions of these errors are provided in Table \ref{error_definition} in the appendix, while Table \ref{tab:error_examples} offers examples for each error label. 

In both settings, the majority of errors stem from modifications unrelated to gender-neutral rewriting and from instances where the model suggests no changes are necessary to render the input text gender-neutral. Additional examples of errors due to unrelated modifications can be found in Table \ref{tab:error_examples_other_modifications} in the appendix. The few-shot setting, however, does show an improvement in neutral rewriting errors (such as POS(part-of-speech) errors and \emph{them} being rewritten as \emph{themselves}) when compared to the zero-shot setting. 

Tables \ref{tab:zero_shot_error_counts}  presents the error distribution for each of the four languages. Upon closer examination of the Finnish data, which has the highest error rate, we found that the errors are primarily due to the longer input length. This increases the scope for modifications of the text that are unrelated to gender-neutral rewriting.

\section{Prompting Templates}
\label{sec:prompting_templates}
\subsection {GPT-3.5 Turbo Prompts }
Figures \ref{fig:zero-shot} and \ref{fig:few-shot-turbo} show the GPT-3.5 Turbo zero-shot and few-shot prompts used in the gender-neutral rewriting task. We use the same prompt across all the language pairs as the task is source agnostic. 

\definecolor{lightgray}{RGB}{240,240,240} %
\renewcommand{\sfdefault}{cmss} 
\begin{figure*}
\begin{tcolorbox}[
    enhanced,
    colback=lightgray,
    colframe=black, 
    boxrule=0.2mm, 
    top=1mm, 
    bottom=1mm, 
    left=1mm, 
    right=1mm,
    fontupper=\sffamily\small,
    ]    
\begin{small}
\noindent
Change all gendered pronouns to use singular "they" instead. Don't modify anything else : \{input\_text\}
\noindent
\end{small}
\end{tcolorbox}
\caption{Zero-shot prompt template utilized in GPT-3.5 Turbo experiments.}
\label{fig:zero-shot}
\end{figure*}
\definecolor{lightgray}{RGB}{240,240,240} %
\renewcommand{\sfdefault}{cmss} 
\begin{figure*}
    
\begin{tcolorbox}[
    enhanced,
    colback=lightgray,
    colframe=black, 
    boxrule=0.2mm, 
    top=1mm, 
    bottom=1mm, 
    left=1mm, 
    right=1mm,
    fontupper=\sffamily\small,
    ]    
\begin{small}
\noindent
Change all gendered pronouns to use singular "they" instead. Don't modify anything else.\\

input : His bike is better than mine.\\
gender neutral variant : Their bike is better than mine.\\\\
input : Jack bores me with stories about her trip.\\
gender neutral variant: Jack bores me with stories about their trip.\\\\
input : He kissed him goodbye and left, never to be seen again.\\
gender neutral variant : They kissed them goodbye and left, never to be seen again.\\\\
input : Is she your teacher?\\
gender neutral variant : Are they your teacher?\\\\
input : Anime director Satoshi Kon died of pancreatic cancer on August 24, 2010, shortly before her 47th birthday.\\
gender neutral variant : Anime director Satoshi Kon died of pancreatic cancer on August 24, 2010, shortly before their 47th birthday.\\\\
input : \{input\_text\}\\
gender neutral variant :

\noindent
\end{small}
\end{tcolorbox}
\caption{Few-shot prompt template utilized in GPT-3.5 Turbo experiments.}
\label{fig:few-shot-turbo}
\end{figure*}
\definecolor{lightgray}{RGB}{240,240,240} %
\renewcommand{\sfdefault}{cmss} 
\begin{figure*}
    
\begin{tcolorbox}[
    enhanced,
    colback=lightgray,
    colframe=black, 
    boxrule=0.2mm, 
    top=1mm, 
    bottom=1mm, 
    left=1mm, 
    right=1mm,
    fontupper=\sffamily\small,
    ]    
\begin{small}
\noindent
I need help with a linguistic annotation task for a translation. I will give you an Turkish sentence along with its translation into English. I would like you to help me find Arbitrarily Gender-Marked Entities (AGMEs), where someone is mentioned without any marked gender in the Turkish sentence, but in the translation they have gender marking. Please follow the following steps:\\\\
1. Identify all unique individuals mentioned in the English translation in the third person and find all words that explicitly indicate those individuals' genders. \\
\hspace*{0.5cm}    - Group words for each individual separately, considering possessive determiners (e.g., "his", "her") as referring to a separate individual from the one indicated by the noun they modify. For example, in "his uncle," "his" and "uncle" refer to two separate individuals.\\
\hspace*{0.5cm}    - Pay attention to gender indicated by kinship terms and other gendered nouns, like "mother", "nephew", "actress".\\ 
\hspace*{0.5cm}    - If the gender is explicitly indicated by pronouns in the target language, consider that gender information for the analysis. (i.e. "she", "he", "him", "her", "his", "hers", "himself", "herself" all explicitly indicate gender)\\
\hspace*{0.5cm}    - Treat names as if they do not indicate a gender, even if they are often associated with a gender. For example, "Michael" could be either male or female, so it does not mark gender.\\
\hspace*{0.5cm}    - Pay attention to how forms or "to be" (particularly "is") can join two mentions of the same individual. For example, in "She is my daughter," "daughter" and "she" refer to the same person.\\\\
2. Find all words in the Turkish source sentence that refer to each of the individuals found in step one.\\\\
3. For each individual, do any of the corresponding words in the Turkish source explicitly indicate a gender. \\
\hspace*{0.5cm}    - Remember, pay attention to gender indicated by kinship words. For example, words like "erkek", "kız" , "amca", "anne" all explicitly indicate gender.\\
\hspace*{0.5cm}    - Remember that some kinship words in Turkish are gender-neutral, such as yeğen. Do not include these as marking gender.\\
\hspace*{0.5cm}    - Treat names as if they do not (e.g. 'Michael' can refer equally well to a man or woman). \\\\
4. Identify any instances where the gender-neutral terms in Turkish have been translated into gender-specific terms in English (AGMEs). \\
\hspace*{0.5cm}    - Answer separately for each individual identified.\\\\
5. Next create a set of variant translations with the following notes:\\
\hspace*{0.5cm}    - If no changes are needed, then just use the original translation exactly as it is.\\
\hspace*{0.5cm}    - Remember to only change the words referring to AGMEs. \\
\hspace*{0.5cm}    - if any gendered words refer to non-AGMEs, leave them untouched.\\
\hspace*{0.5cm}    - Do not make assumptions about heterosexual relationships. Men can have husbands and boyfriends. Women can have wives and girlfriends.\\\\
 
    Please create these three variant translations:\\
\hspace*{0.5cm}  a. If any individuals are AGMEs and are referred to with gendered words in English, rewrite the English translation changing only those words to use their gender-neutral variants where possible. Use singular "they" instead of he, she, etc. Use "themselves" for gender neutral singular reflexives (never "themself"). Change nothing else.\\
\hspace*{0.5cm}  b. rewrite the English translation so that any masculine words referring to AGMEs are replaced by their feminine variants. Don't change any words referring to non-AGMEs. Change nothing else.\\
\hspace*{0.5cm}  c. rewrite the English translation so that any feminine words referring to AGMEs are replaced by their masculine variants. Don't change any words referring to non-AGMEs. Change nothing else.\\\\

\noindent
\end{small}
\end{tcolorbox}
\caption{Part I of Few-shot prompt template utilized in GPT-4 experiments.}
\label{fig:few-shot-gpt-4-part-1}
\end{figure*}
\definecolor{lightgray}{RGB}{240,240,240} %
\renewcommand{\sfdefault}{cmss} 
\begin{figure*}
    
\begin{tcolorbox}[
    enhanced,
    colback=lightgray,
    colframe=black, 
    boxrule=0.2mm, 
    top=1mm, 
    bottom=1mm, 
    left=1mm, 
    right=1mm,
    fontupper=\sffamily\small,
    ]    
\begin{small}
\noindent 
Example 1 -\\
Source Sentence: Amcası kendi kendine konuşuyor.\\
Original Translation: His uncle talks to himself.\\
1. individual 1: "His" is masculine.
   individual 2: "uncle", "himself" are masculine.\\
2. individual 1: no explicit words in the source
   indiviudal 2: "Amcası", "kendi kendine"\\
3. individual 1: no words indicate gender
   indivudual 2: "Amcası" is masculine.\\
4. individual 1: AGME - masculine in translation ("His"), but gender neutral in the source (no explicit words)
   individual 2: not an AGME - gender is masculine in both the source ("Amcası") and translation ("uncle")\\
5. a. Their uncle talks to himself.\\
\hspace*{0.3cm} b. Her uncle talks to himself.\\
\hspace*{0.3cm}  c. His uncle talks to himself.\\\\
 
Example 2 -\\
Source Sentence: Annem öğle yemeğini yalnız yiyordu.\\
Original Translation: My mother ate her lunch alone.\\
1. individual 1: "mother", "her" are feminine.\\
2. individual 1: "Annem" \\
3. individual 1: "Annem" is feminine.\\
4. individual 1: not an AGME since gender is feminine in both the source and translation\\
5. None\\\\
 
Example 3 -\\
Source Sentence: O benim kızım\\
Original Translation: She is my daughter.\\
1. individual 1: "she", "daughter" are feminine.\\
2. individual 1: "O", "kızım"\\
3. individual 1: "kızım" is feminine.\\
4. individual 1: not an AGME since gender is feminine in both the source and translation\\
5. None\\
 
Source Sentence: \{source\_sentence\}\\
Original Translation: \{original\_translation\}\\

\noindent
\end{small}
\end{tcolorbox}
\caption{Part II of Few-shot prompt template utilized in GPT-4 experiments.}
\label{fig:few-shot-gpt-4-part-2}
\end{figure*}

\subsection {GPT-4 Prompts}
The full prompt for Turkish-English can be found in Figures \ref{fig:few-shot-gpt-4-part-1} and \ref{fig:few-shot-gpt-4-part-2}. Prompts for other languages use the same structure, but examples are customized to fit those languages.

\section{Rewriting with GPT-4}

Our solution uses chain-of-thought prompting \citep{chain-of-thought-wang-etal} to elicit GPT-4 to produce three variant translations for each input source-translation pair -- all-neutral, all-female and all-male, while leaving any gendered words associated with non-AGMEs unmodified. We ask it to work step-by-step through the process of identifying AGMEs before finally rewriting the original translation:

\begin{itemize}
    \item Identify unique individuals mentioned in the target, as well as any gendered words that refer to them.
    \item Identify words in the source that refer to those same individuals.
    \item Determine which source words mark for gender.
    \item Designate any individuals referred to by gendered words in the target, but not in the source as AGMEs
    \item Produce a neutral, feminine and masculine variant translation where any gendered words referring to AGMEs are modified to match the respective gender.
\end{itemize}

Each step in the prompt is accompanied by detailed clarifications and example vocabulary. The prompt also includes three full examples, customized per source language. The examples indicate that "None" should be returned in lieu of translation variants when there are no AGMEs present. The full prompt for Turkish-English is shown found in Figures \ref{fig:few-shot-gpt-4-part-1} and \ref{fig:few-shot-gpt-4-part-2}.

\section{Mitigation Strategies Based on Source}

Rather than rewriting English-target translations into feminine, masculine, and neutral forms, one could use the source sentence as input to create these three variants directly. This section explains how GATE X-E can be employed to assess such a system. 

The first step is to verify that the generated feminine, masculine, and neutral variants are the same, except for changes related to gender. This is a crucial step as it ensures that the meaning of the translation remains consistent, regardless of the gender. If there are differences in the translations beyond the gender-related changes, it could imply that the translation is not accurate or is introducing additional bias. After this, the generated output can be compared with the feminine, masculine, and neutral references provided in GATE X-E using contextual MT evaluation metrics.

\citet{kuczmarski2018} initially explored a source-based debiasing approach in which they enhanced a Neural Machine Translation (NMT) system to produce gender-specific translations. This was achieved by adding an additional input token at the beginning of the sentence to specify the required gender for translation (e.g., \emph{<2FEMALE> O bir doktor $\rightarrow$ She is a doctor}). However, they encountered challenges in generating masculine and feminine translations that were exactly equivalent, with the exception of gender-related changes. As a result, they later switched to a target-based rewriting approach in their subsequent work \citep{johnson2020}.

\begin{table*}
  \centering
  \begin{tabular}{p{5cm} p{5cm} p{5cm}}
    \toprule
    Original & Gender-Neutral & Gendered Alternatives \\
    \midrule
    The teacher compared my poem with one of \textcolor{purple}{his}.
 & The teacher compared my poem with one of \textcolor{purple}{theirs}.
 & The teacher compared my poem with one of \textcolor{purple}{hers.}\\
 
    &  & The teacher compared my poem with one of \textcolor{purple}{his}.
\\
    \bottomrule
  \end{tabular}
\caption{Examples illustrating the generation of gendered alternatives using gender-neutral rewrites }
\label{tab:original_neutral_gendered}
\end{table*}

\begin{table*}
\centering
\begin{tabular}{|p{4cm}|l|p{8cm}|}
\hline
\textbf{Error Category} & \textbf{Error Label} & \textbf{Description} \\ \hline
\multirow{2}{=}{\parbox{4cm}{Corrections unrelated to neutral rewriting}} & Comma & Comma added or removed. \\ \cline{2-3}
 & Other corrections & Error corrections that are not related to making the text gender-neutral. \\ \hline
\multirow{4}{=}{\parbox{4cm}{Neutral Rewriting}} & POS & Wrong form of ‘they’ produced by the rewriter. \\ \cline{2-3}
 & SVA & Failure to make correct subject-verb agreement. \\ \cline{2-3}
 & Them $\rightarrow$ Themselves & Second appearance of them changed to themselves. \\ \cline{2-3}
 & 'None' response & The rewriter returns ‘None’ suggesting no changes needed to make the input gender neutral. \\ \hline
Irrelevant modifications & Other modifications & Modifications unrelated to neutral rewriting. \\ \hline
\end{tabular}
\caption{Description of the error labels}
\label{error_definition}
\end{table*}

\begin{table*}
    \centering
    \begin{tabular}{|c|p{3cm}|p{3cm}|p{4cm}|}
        \hline
        \textbf{Error Label} & \textbf{Input} & \textbf{Output} & \textbf{Reference} \\
        \hline
        Comma & Well, you surprised me!\textbf{,} Afshin said as she opened the door and saw Mary standing there. & Well, you surprised me! Afshin said as they opened the door and saw Mary standing there. & Well, you surprised me!\textbf{,} Afshin said as they opened the door and saw Mary standing there.
 \\
        \hline
        Other corrections & I \textbf{have} never heard of him before that.
 & I \textbf{had} never heard of them before that.
 & I \textbf{have} never heard of them before that. \\
        \hline
POS & The secretary noted down what \textbf{her} boss had said. & The secretary noted down what \textbf{they} boss had said.
 & The secretary noted down what \textbf{their} boss had said. \\
        \hline
SVA & \textbf{Does} she come here every week? & \textbf{Does} they come here every week?
 & \textbf{Do} they come here every week? \\
        \hline
Them $\rightarrow$ Themselves & She saw \textbf{her} play baseball. & They saw \textbf{themselves} play baseball.
 & They saw \textbf{them} play baseball. \\
        \hline
'None' response & He has no capacity to be a teacher. & none & They have no capacity to be a teacher. \\
        \hline
Other modifications& In any case, I will tell him about the critical tone \textbf{your} House has adopted on this issue. & In any case, I will tell them about the critical tone \textbf{their} House has adopted on this issue.
 & In any case, I will tell them about the critical tone \textbf{your} House has adopted on this issue. \\
        \hline
    \end{tabular}
    \caption{Examples for the error labels described in Table \ref{error_definition}}
    \label{tab:error_examples}
\end{table*}

\begin{table*}
    \centering
    \begin{tabular}{|p{7cm}|p{7cm}|}
        \hline
        \textbf{Output} & \textbf{Reference} \\
        \hline
They advised them to give up smoking, but they wouldn't listen.
 & They advised them to give up smoking, but they wouldn't listen \textcolor{red}{to them}.
 \\
        \hline
\textcolor{red}{They} w\textcolor{red}{ere} able to hold back their anger and avoid a fight.
 & \textcolor{red}{Jim} w\textcolor{red}{as} able to hold back their anger and avoid a fight.
\\
        \hline
The news that they had got\textcolor{red}{ten} injured was a shock to them.
 & The news that they had got injured was a shock to them.
 \\
        \hline
They have done it with the\textcolor{red}{ir} colleagues and the Committee of Legal Affairs.
 & They have done it with the colleagues and the Committee of Legal Affairs.
 \\
        \hline
In this respect \textcolor{red}{,} they have been very successful.
 &In this respect \textcolor{red}{I believe that} they have been very successful.
 \\
        \hline
They cannot be older than \textcolor{red}{me}.
 & They cannot be older than \textcolor{red}{I}.
 \\
        \hline
They suggested \textcolor{red}{goin}g to the theater, but there weren't any performances that night.
 & They suggested \textcolor{red}{to} g\textcolor{red}{o} to the theater, but there weren't any performances that night.
 \\
        \hline
    \end{tabular}
    \caption{More examples of errors of type 'Other Modifications'. Differences are in \textcolor{red}{red}.}
    \label{tab:error_examples_other_modifications}
\end{table*}

\begin{table*}
\centering
\small
\begin{tabular}{llcccccccccc}
\toprule
    {\bf  Category} & {\bf Error Label} & \multicolumn{4}{c}{\bf Zero-shot  } & \multicolumn{4}{c}{\bf Few-shot  }\\
    & & {\bf tu } & {\bf hu  } & {\bf  fi} & {\bf  fa } & {\bf tu } & {\bf hu  } & {\bf  fi} & {\bf  fa }\\
\midrule
\midrule
    \multirow{2}{*}{Corrections} &  Comma & 0 &  0  & 2 & 2 & 0 &  0  & 0 & 0  \\
    & Other Corrections & 0 &  0 &  0 & 2 & 0 &  0 &  0 & 0 \\
    \midrule
    \multirow{4}{*}{Neutral rewriting} & POS  & 2  &  1  &  9  & 4 & 0  &  0  &  0  & 1 \\
     &  SVA
 & 5 &  5 &  3 & 0  & 0 &  9 &  0 & 2 \\
    & Them $\rightarrow$ Themselves
 &  0 &  10 &  10 & 0 &  0 &  4 &  0 & 0\\
     & 'None' response
 &  4 &  6 &  20 & 11 &  4 &  6 &  22 & 10 \\
        \midrule
    \multirow{1}{*}{Irrelevant Modifications} & 
      Other modifications  & 16 &  16  &   33  & 8 & 4 &  10  &   32  & 10 \\
         \midrule
    \multirow{1}{*}{Total} & 
      & 27 &  38  &   77  & 27 
      & 4 &  19  &   54  & 23\\   
\bottomrule
\end{tabular}
\caption{Error analysis of GPT-3.5 Turbo's zero-shot and few-shot performance in English gender-neutral rewriting task.}
\label{tab:zero_shot_error_counts}
\end{table*}
\newpage

\begin{table*}
\centering
\begin{tabular}{llll}
\toprule
Category & Feminine & Masculine & Neutral \\
\midrule
Subject & She & He & They \\
Object & Her & Him & Them \\
Possessive Determiner & Her & His & Their \\
Possessive Pronoun & Hers & His & Theirs \\
Reflexive & Herself & Himself & Themselves \\
\bottomrule
\end{tabular}
\caption{Pronoun categories}
\label{tab:pronoun_cats}
\end{table*}

\begin{table*}
\centering

\begin{tabular}{|lll|}
\hline
he	& she	& him \\
her	& his	& himself \\
herself	& Ms	& Mrs \\
Ms.	& Mrs.	& madam \\
woman	& women	& actress \\
actresses	& airwoman	& airwomen \\
aunts	& aunt	& uncle \\
uncles	& brother	& brothers \\
boyfriend	& boyfriends	& girlfriend \\
girlfriends	& girl	& girls \\
bride	& brides	& sister \\
sisters	& businesswoman	& businesswomen \\
chairwoman	& chairwomen	& chick \\
chicks	& mom	& moms \\
mommy	& mommies	& grandmother \\
daughter	& daughters	& mother \\
mothers	& female	& females \\
gal	& gals	& lady \\
ladies	& granddaughter	& granddaughters \\
grandmother	& grandmothers	& grandson \\
grandsons	& grandfather	& grandfathers \\
wife	& wives	& queen \\
 queens	& policewoman	& policewomen \\
princess	& 	princesses & spokeswoman \\
 spokeswomen	& stepson	&  stepdaughter \\
  stepfather	& stepmother	&  stepgrandmother \\
    stepgrandfather	& 	&   \\
   \hline
\end{tabular}
\caption{Gendered English Nouns and Pronouns}
\label{tab:gendered_noun_pronouns}
\end{table*}

\end{document}